%% file: arXiv_Submission.tex
\documentclass[10pt,twocolumn,letterpaper]{article}

\usepackage{wacv}              % To produce the CAMERA-READY version
\usepackage{url}
\usepackage{graphicx}
\usepackage{amsmath}
\usepackage{amssymb}
\usepackage{multirow}
\usepackage{adjustbox}
\usepackage{xcolor}
\usepackage{booktabs}
\usepackage{tikz}
\usepackage{pgfplots}
\usepackage{pgfplotstable}
\usepackage{listings}
\usepackage{caption}
\usepackage{subcaption}
\usepackage{bbm}
\usepackage{amssymb}% http://ctan.org/pkg/amssymb
\usepackage{pifont}% http://ctan.org/pkg/pifont
\newcommand{\cmark}{\ding{51}}%
\DeclareMathOperator*{\argminA}{arg\,min}
\DeclareMathOperator*{\argmaxA}{arg\,max}

\definecolor{red1}{HTML}{EA484B}
\definecolor{red-light}{HTML}{f4a4a5}
\definecolor{yellow1}{HTML}{FDAE61}
\definecolor{yellow-light}{HTML}{fdcc9b}
\definecolor{green1}{HTML}{19c20a}
\definecolor{green-light}{HTML}{a6fa9e}
\definecolor{blue1}{HTML}{2B83BA}
\definecolor{blue-light}{HTML}{acd3ec}
\definecolor{orange1}{HTML}{ff9900}
\definecolor{purple1}{HTML}{ad33ff}
\definecolor{purple-light}{HTML}{d699ff}

\usepackage[pagebackref,breaklinks,colorlinks]{hyperref}

\usepackage[capitalize]{cleveref}
\crefname{section}{Sec.}{Secs.}
\Crefname{section}{Section}{Sections}
\Crefname{table}{Table}{Tables}
\crefname{table}{Tab.}{Tabs.}

\def\wacvPaperID{2133} % *** Enter the WACV Paper ID here
\def\confName{WACV}
\def\confYear{2024}

\begin{document}

\title{pSTarC: Pseudo Source Guided Target Clustering \\ for Fully Test-Time Adaptation} 

% \and

\author{
Manogna Sreenivas$^\dagger$, Goirik Chakrabarty$^*$, Soma Biswas$^\dagger$  \\
$^\dagger$IISc Bangalore \quad $^*$IISER Pune\\
% % IISc, Bangalore \quad IISER, Pune\\
% $^\dagger${\tt\small{\{manognas, somabiswas\}}@iisc.ac.in}  \quad
% $^*${\tt\small goirik.chakrabarty@students.iiserpune.ac.in} \\
% Manogna Sreenivas \\
% IISc, Bangalore\\
% \and
% Manogna Sreenivas \\
% IISc, Bangalore\\
% % First line of institution2 address\\
% {\tt\small {\{manognas, somabiswas\}}@iisc.ac.in}
}

% \and

\maketitle

%%%%%%%%% ABSTRACT
%%%%%%%%% ABSTRACT
\begin{abstract}
% Test Time Adaptation (TTA) of deep neural networks is an important paradigm in machine learning as it enables the model to adapt in real-world test scenarios, where the distribution of data may differ from the distribution of the training data. 
% In this work, we address the challenging task of fully TTA on real-world domain shift data, which is relatively less explored.
% %\color{red} We observe that existing fully TTA approaches do not always improve the performance on real-world domain shift data. \color{black}
% Inspired by the effectiveness of target clustering techniques, we propose a simple, yet effective framework that leverages the source classifier to generate pseudo-source samples.
% We show that effectively generating these samples and choosing them as neighbors of each target sample can help to align and cluster the targets close to the source, which in turn improves the TTA performance.
% The proposed \textbf{p}seudo \textbf{S}ource guided \textbf{Tar}get \textbf{C}lustering (\textbf{pSTarC}) framework operates in the fully test-time adaptation protocol, i.e. does not assume any access to the source data.
% Extensive experiments across three real-world domain shift datasets, namely VisDA~\cite{visda}, Office-Home~\cite{office-home}, DomainNet~\cite{moment} and the corruption benchmark CIFAR-100C demonstrates the effectiveness of pSTarC, both in terms of performance and computational requirements.

Test Time Adaptation (TTA) is a pivotal concept in machine learning, enabling models to perform well in real-world scenarios, where test data distribution differs from training. In this work, we propose a novel approach called \textbf{p}seudo \textbf{S}ource guided \textbf{Tar}get \textbf{C}lustering (\textbf{pSTarC}) addressing the relatively unexplored area of TTA under real-world domain shifts. This method draws inspiration from target clustering techniques and exploits the source classifier for generating pseudo-source samples. The test samples are strategically aligned with these pseudo-source samples, facilitating their clustering and thereby enhancing TTA performance. pSTarC operates solely within the fully test-time adaptation protocol, removing the need for actual source data.
Experimental validation on a variety of domain shift datasets, namely VisDA, Office-Home, DomainNet-126, CIFAR-100C verifies pSTarC's effectiveness. This method exhibits significant improvements in prediction accuracy along with efficient computational requirements. Furthermore, we also demonstrate the universality of the pSTarC framework by showing its effectiveness for the continuous TTA framework. The source code for our method is available at \href{https://manogna-s.github.io/pstarc}{https://manogna-s.github.io/pstarc}

\end{abstract}

\section{Introduction}
% Domain adaptation problems have evolved over time as researchers have attempted to address various challenges and limitations in the field of machine learning.

% The success trajectory of deep networks has been ever growing over the past decade. This can be attributed to the curation of large datasets, tremendous improvements in computing power and advances in algorithms and architectures. This impressive performance is achieved primarily when the test data is from the same distribution as that of the training data. In reality, however, this is a very hard assumption. A classic example where this assumption breaks is that of autonomous vehicles. Suppose, the deployed deep frameworks have been trained and tested on data collected in clear weather conditions. In reality, these vehicles can witness rainy, snowy or even adverse conditions like sandstorms. While we humans continue to perform well on encountering new scenarios, the performance of deep networks significantly degrade. This being highly undesirable, recently, there has been increasing interest in studying and improving the robustness of deep networks.

Over the past decade, deep networks have shown a continuous upward trend due to the availability of large datasets~\cite{imagenet, mscoco, pascal-voc}, significant improvements in computing power, and advancements in algorithms~\cite{faster_rcnn, mask_rcnn} and architectures~\cite{vgg, resnet}. But while humans can adapt seamlessly to new domains, the performance of deep networks deteriorate significantly when the test  and training distributions differ.  
In practical scenarios, a trained model is often deployed in an unseen test environment, so equipping it with good adaptation capabilities to mitigate the adverse effects of any domain shift is crucial. 
Additionally, since access to the source data may be difficult because of privacy concerns or storage limitations, 
there is a significant interest in the following research directions: 
(i) Source-free Domain Adaptation (SFDA)~\cite{shot, nrc, aad}, which assumes access to the source model and a large amount of unlabeled test data and 
(ii) Test-Time adaptation (TTA)~\cite{tent, lame, adacontrast}, where test data arrives in an online manner, one batch at a time, allowing for one-step model adaptation followed by prediction.
SFDA and TTA methods have been developed independently, resulting in fundamentally different approaches.

Here, we address the challenging and more practical task of swiftly adapting models without the need for extensive accumulation of test data, i.e. the TTA setting.
Unlike SFDA methods which have been evaluated on real world domain shift datasets like VisDA~\cite{visda}, DomainNet~\cite{moment} and Office-Home~\cite{office-home}, TTA methods have primarily been evaluated within the confines of artificially corrupted data. 
It is only recently that researchers have started to address the TTA task for such real-world domain shifts~\cite{adacontrast,src_proxy_icml23,csfda}.

In this work, we propose a simple yet effective TTA strategy termed {\bf p}seudo {\bf S}ource guided {\bf Tar}get \textbf{C}lustering (pSTarC). 
It is inspired by the exceptional performance of SFDA techniques like SHOT \cite{shot}, NRC \cite{nrc}, and AaD \cite{aad} in the context of the real world domain shift benchmarks. Notably, contemporary SFDA methods, including NRC and AaD, concentrate on refining target sample clustering, leveraging the luxury of abundant unlabeled target data. To extend this SFDA principle to TTA, one compelling avenue is the maintenance of a feature bank, which dynamically populates as new target data becomes available, enriching the adaptation process. While approaches like AdaContrast~\cite{adacontrast} have successfully harnessed this concept for TTA, they need to store auxiliary components like the momentum encoder and key features, a constraint that might not align well with an online TTA framework.

Our proposed pSTarC approach aims to leverage the power of SFDA objectives while adhering to the principle of minimizing memory and storage requirements for TTA task. Generally, the source-trained classifier remains unchanged during TTA to preserve the valuable class-discriminative insights gained from the source. Building on this insight, we introduce a novel strategy: utilizing the classifier to generate a diverse array of pseudo-source samples, thereby steering the target clustering process. Impressively, our findings reveal that generating as few as 20 pseudo-source samples per class is adequate to achieve state-of-the-art TTA performance, without imposing a significant burden on storage demands.
Thus, the main contributions of this work can be summarized as follows: 
\begin{enumerate}
    \item We propose the pSTarC framework, which generates pseudo-source samples to guide the target clustering during test time adaptation.
    % \color{blue}
    \item We strive to achieve TTA using SFDA objectives, which not only improves the TTA performance significantly for real domain-shifts, but also helps to unify the seemingly disparate research directions. %(ii) facilitates hybrid adaptation, i.e. adapt (SFTA) with some amount of data and then activate the TTA mode.
    % \color{black}
    \item pSTarC outperforms the state-of-the-art TTA techniques on Office-Home and DomainNet, and at par on VisDA, while requiring much lesser memory. 
    \item pSTarC also seamlessly works in Continual Test-Time Adaptation (CTTA)~\cite{cotta} scenario, where the test distribution changes with time. Here, its performance is at par with current state-of-the-art approaches on the large-scale DomainNet-126 benchmark. 
\end{enumerate}   
In a nutshell, pSTarC aligns seamlessly with our objective to pave the way for swift, efficient TTA in the face of real-world domain shifts, building upon the insights garnered from the relationship between SFDA techniques and such demanding benchmarks.

%The rest of the paper is organized as follows. 
%Section~\ref{relatedworks} discusses the related work in literature.
%The problem definition and proposed approach is described in Section~\ref{problem} and~\ref{main} respectively.
%The experimental results are described in Section~\ref{expt} followed by further analysis and a brief conclusion.

% *****************************

\section{Related Works}
\label{relatedworks}
Here, we discuss the related work in Source-Free Domain Adaptation, Test-Time Adaptation, Continuous Test Time Adaptation and  Model Inversion. \\

\noindent\textbf{Source-free domain adaptation} (SFDA) aims to adapt a source domain trained model to a target domain without access to any labeled data from either the source or target domain. SFDA methods typically assume access to abundant unlabeled data from the target domain 
and leverage the structure of the data to refine the target predictions.
~\cite{shot} proposes to cluster target features by mutual entropy maximization along with pseudo labeling, while keeping the classifier fixed. ~\cite{adacontrast} extends the idea in ~\cite{shot} proposing to refine the pseudo labels using a feature bank, alongside doing self-supervised contrastive learning~\cite{simclr}. 
Another line of work include~\cite{nrc, aad}, where they exploit the inherent semantic structure of the target features extracted from the source model. They reinforce consistency between the predictions of a sample and its local neighbors while also ensuring diversity to avoid degenerate predictions. 

\noindent\textbf{Test Time Adaptation} (TTA) further relaxes the assumptions on data availability compared to SFDA. 
TENT~\cite{tent} proposed the more practical fully test time adaptation setting, where source data cannot be accessed at all, and the model can only utilize the test samples in each batch encountered in an online manner for adaptation. 
They propose minimizing the entropy of the model predictions on the test data. 
%More recent methods such as LAME \cite{lame} and AdaContrast \cite{adacontrast} use different approaches for the FTTA setting. 
More recently LAME~\cite{lame} uses Concave-Convex Procedure (CCCP) to modify the feature vectors to obtain better classification, while AdaContrast~\cite{adacontrast} addresses SFDA and TTA settings by using contrastive learning with nearest neighbour soft voting for online pseudo label refinement. C-SFDA~\cite{csfda} uses curriculum learning in a Teacher Student framework.
Other works like EATA \cite{eata} uses a small buffer from source distribution. TTN \cite{ttn} trains a modified BN layer to leverage source data for improved TTA. In \cite{src_proxy_icml23}, they synthesize source proxy images by condensing the source dataset, which is then used during TTA after stylizing them to match the test distribution.
Our work falls in the category of fully test-time adaptation~\cite{tent, adacontrast, csfda, cotta}. 

\noindent\textbf{Continual Test Time Adaptation} (CTTA) As a further extension of TTA, the concept of continual test-time adaptation (CTTA) has been recently introduced\cite{cotta}. This protocol recognizes the dynamic nature of the testing environment, where the test domain evolves over time. CoTTA~\cite{cotta} adopts strategies like weight-averaged and augmentation-averaged predictions in a teacher-student framework to mitigate error accumulation. Additionally, it retains a fraction of neurons with source pre-trained weights during each iteration to prevent catastrophic forgetting, thus enabling model adaptation while preserving source knowledge. RMT~\cite{rmt} is a recent CTTA method that uses symmetric cross-entropy loss and contrastive loss in a teacher student framework.

\noindent\textbf{Model inversion} is a recent research direction explored in ~\cite{imagine, naturalinversion, magic} for image generation where they optimize the
% Our idea is inspired from the model inversion techniques~\cite{imagine, naturalinversion, magic} for image generation. 
% Model inversion methods involve modifying the 
input space to generate an image $\hat{x}$ using a pre-trained deep network. To do this, given an arbitrary target $y$ which can be a label or a reference image, a trainable input $\hat{x}$ in the image space is initialized with random noise. This input space is then optimized by minimizing a loss function $ \mathcal{L}(\hat{x}, y)$, which is usually cross-entropy loss and a regularizer $\mathcal{R}(\hat{x})$ to induce  natural image prior.  
The training is done in an adversarial manner by alternating between the optimization of the synthesized image and that of the discriminator weights. 
% Such methods are also used during training to improve the model robustness against model inversion attacks~\cite{modelinv} or adversarial attacks.  
% As we have no access to the source data during, test time,
Inspired by the effectiveness of these methods, here we propose a classifier guided {\em feature generation} approach, which is used for generating pseudo-source samples for guiding the clustering of the target data.

\section{Problem Setting \& Motivation}
\label{problem}
Firstly, the source model is trained using labeled source data. Then, in the Test Time Adaptation (TTA) stage, this model is adapted using the test batches in an online manner.

% Test Time Adaptation (TTA) setting usually consists of a source training phase (which cannot be controlled), and the TTA stage, where the trained model in the previous stage is adapted using the test batches in an online manner. 

\noindent\textbf{Source training:} 
The model is first trained using labeled source data $\mathcal{D}_{s} = \{x_{i}^{s}, y_{i}^{s}\}_{i=1}^{n_s}$ comprising of $C$ classes. Here, $x_{i}^{s} \in \mathcal{X}_{s}$ and $y_{i}^{s} \in \mathcal{Y}_{s}$ denote the source sample and its class label, and $n_{s}$ is the number of training samples. We denote the source model as $\mathbf{F_{s}= H_{s} \circ G_{s}}$, where $\mathbf{G_{s}}$ is the feature extractor and $\mathbf{H_{s}}$ is the classifier. 
Following ~\cite{shot, aad, adacontrast}, the source model $\mathbf{F_{s}}:\mathcal{X}_{s} \rightarrow \mathcal{Y}_{s} $ is trained by minimizing the label-smoothing cross entropy loss as 
\begin{displaymath}
 \mathcal{L}_{s r c}\left(f_s ; \mathcal{D}_s\right)=  -\mathbb{E}_{\left(x^s, y^s\right) \in \mathcal{D}_s} \sum_{c=1}^C \Tilde{y}_c^{s} \log p_{c}, %\delta_k\left(f_s\left(x_s\right)\right),
\end{displaymath}
% where $\delta_k(a) = \frac{exp(a_{k})}{\sum exp(a_{k})}$
% where $p_{k} = \delta_k\left(f_s\left(x_s\right)\right)$, 
where  $p_{c} = \delta_c\left(f_s\left(x^{s}\right)\right)$ is the softmax score for class $c$, $\delta$ being the softmax function.
The smooth label is computed as $\tilde{y}_c^s=(1-\alpha) y_c^s+\alpha / C$,  where the smoothness coefficient $\alpha$ is set to $0.1$.
\\

% In the second stage of TTA, 
\noindent\textbf{Test Time Adaptation:} 
Given the source model $\mathbf{F_{s}}$, during TTA, the target model $\mathbf{F_t}$ is initialized with the source model $\mathbf{F_s}$. 
We only have access to the unlabeled test samples $x_{t}$ coming in batches from an unseen test distribution $\mathcal{D}_{t}$. Here, we address the closed setting where the source and target samples come from the same $C$ classes. The goal is to continuously adapt $\mathbf{F_{t}}: \mathcal{X}_{t} \rightarrow \mathcal{Y}_{t} $ using the unlabeled samples $x_{t} \in \mathcal{X}_{t}$ (in batches) in an online manner. \\

\noindent\textbf{Continual Test Time Adaptation:}
In addition to the above setup, the test data can come from multiple domains which changes over time such that $D_t^{(1)} \neq D_t^{(2)} \neq \hdots \neq D_s$ leading to the continual test time adaptation scenario.

\color{black}

\begin{figure*}[h]
    \centering
    \includegraphics[width=0.75\textwidth]{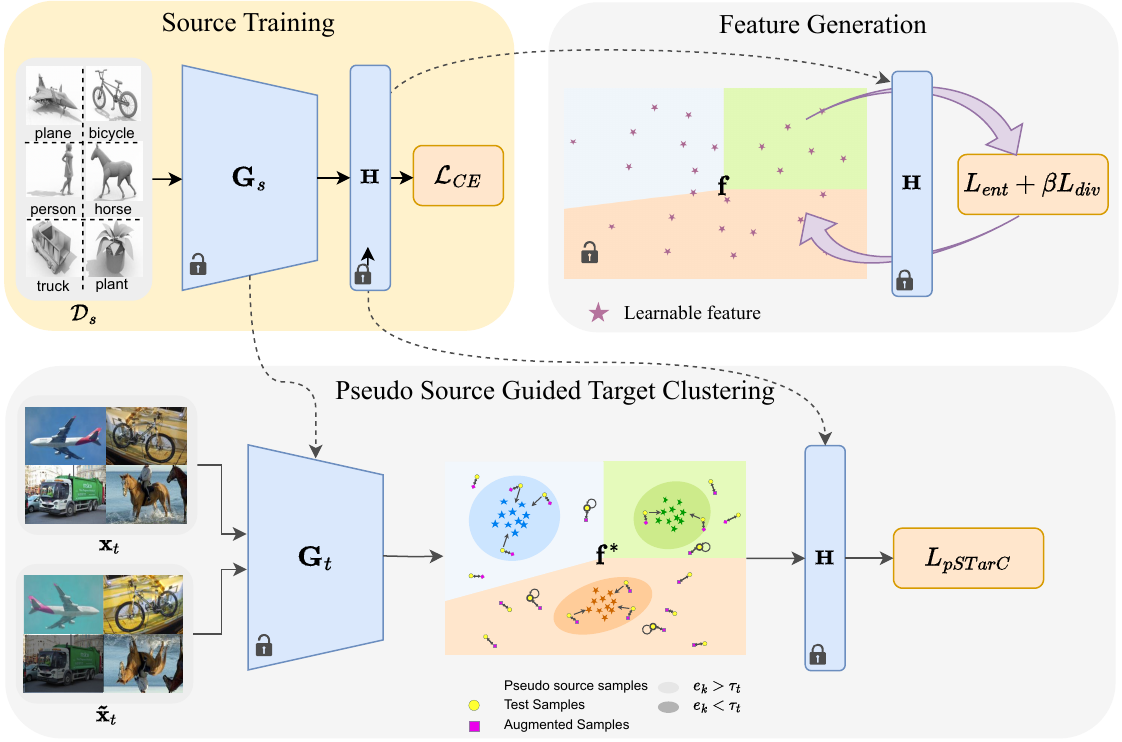}
    \caption{pSTarC Framework: (1) Feature Generation: We randomly initialize a feature bank $\mathbf{f}$ which is iteratively optimized keeping the classifier $\mathbf{H}$ fixed to minimize the entropy of the features while maximizing the diversity across classes using the loss in eqn~(\ref{eq:fea_gen}). (2) Given the learnt features, we aim to bring the low entropy samples towards the corresponding pseudo-source features. We anchor the high entropy target samples to its own prediction. We also enforce consistency between the predictions of the test sample and its strong augmentation.}
    \label{fig:protocols}
\end{figure*}

\section{Proposed Framework} 
\label{main}
The proposed pSTarC framework is based on effectively clustering the target samples which are available during test time. 
Our formulation is inspired by the clustering framework proposed in the state-of-the-art SFDA technique, AaD~\cite{aad}, which we briefly describe below. \\ \\
%, which also helps to achieve our additional objective of developing unified formulations for SFDA and TTA techniques. 
%We briefly describe the AaD approach below: 
% *********************************
\textbf{Attracting and Dispersing (AaD):} 
%This~\cite{aad} is a simple and effective approach recently proposed for SFDA. 
AaD~\cite{aad} treats SFDA as an unsupervised clustering problem, where consistency is enforced between predictions of local neighbourhood features, while also ensuring diversity in the feature space. The test objective for a sample $x_i$ from a test batch $\mathbf{x}_t$ is 
\begin{equation}
% \mathcal{L}_{AaD}(\mathbf{x}_t) = \mathbf{E}_{x_{i}\in \mathbf{x}_{t}} \mathcal{L}(x_{i});
% \qquad \textrm{where} \hspace{0.4cm}   
\mathcal{L}(x_{i}) = -\sum_{x_j \in \mathcal{N}_{i}} p_i^{T} p_j + \lambda \sum_{x_{m}\in \mathbf{x}_{t}} p_i^{T} p_m
\end{equation}
where $p_i$ refers to the softmax prediction vector of the sample $x_{i} \in \mathbf{x}_t$, $p_j$ in the first term corresponds to the prediction vectors in its neighborhood $\mathcal{N}_i$, $p_m$ in the second term corresponds to the prediction vectors of the samples $x_m$ in the current batch $\mathbf{x}_{t}$.

Now, we describe the proposed pSTarC framework for fully TTA task, which we also illustrate in Fig.(~\ref{fig:protocols}).
In a TTA setting, as mentioned before, 
the labeled source samples are unavailable, and only the source model is available for adaptation. 
In addition, since the number of samples in a batch is usually quite low, it is a common practice to freeze the source trained classifier and update only the feature extractor to align target features with those of the source. 
Hence, we set $\mathbf{H_{t} = H_{s} = H}$ and only update the feature extractor $\mathbf{G_{t}}$ using the test data in an online manner.
% Though storing a feature bank is the most straight-forward extension of SFDA to TTA task as done in AdaContrast~\cite{adacontrast}, it requires storing the adapting model to have access to the updated features, which increases the storage cost. 
% In contrast, during adaptation, the source and target classifiers remain unchanged, i.e. $h_{t} = h_{s} = h$, whereas only the feature extractor $g_{t}$ gets adapted to the target data in an online manner.
The goal is to adapt the test features such that they align with the source features so that the classifier $\mathbf{H}$ is transferable to test data.
% which is usually used in UDA frameworks, when both the source and target data are available simultaneously. 
The classifier, being trained in a supervised manner using abundant source data, defines the decision boundaries for which the source data is perfectly classified.
We leverage this fact to synthesize pseudo-source features, which are used to guide the target clustering.
Given the source model, this process is only done once to store few features and corresponding prediction scores, and can be utilized throughout the TTA process.
We describe the feature generation and clustering in detail below.
% ***************************************
\subsection{Pseudo Source Feature Generation}
\label{sec:fea_gen}
Since the decision boundaries in the feature space remain fixed (due to the classifiers remaining unchanged), it is important to align the target features with the original source features, which will inherently lead to better clustering and hence better classification of the target samples.  
First, we utilize the fixed source classifier to synthesize pseudo-source features. %The source-trained classifier acts as a proxy for the source data, and its decision boundaries are preserved in the target domain. We can use this information to generate synthetic source-like features that are aligned with the source domain. 
By aligning the target to these generated features, we hope to improve the adaptation performance of the model and make it more robust to the domain shift between the source and target domains.

Here, we aim to generate, say $N$ pseudo-source features, where $N = C \times n_c$, $C$ being the number of classes and $n_c$ is the number of samples per class. We first randomly initialize a feature bank $\mathbf{f} \in \mathcal{R}^{N\times d}$, where $d$ is the feature dimension.
% Here, we aim to generate, say  $n$ features per class, for which we first randomly initialize a feature bank $\mathbf{f} \in \mathcal{R}^{N\times d}$, where $d$ is the feature dimension and $N = C \times n$, $C$ being the number of classes.
To compute the pseudo-source features, we use the information maximization loss which is a combination of entropy minimization and diversity maximization. 
These losses have been widely used in unsupervised clustering methods~\cite{shot} to optimize a feature extractor to make the predictions of unlabeled samples diverse and confident. 
However, our objectives are very different. While they aim to learn a good feature extractor, our goal is to synthesize pseudo-source features given the source trained classifier $\mathbf{H}$.  
% The entropy loss ensures that the feature vectors have high confidence in their predictions. 
We want to generate features which are likely to be correctly classified by the source classifier. 
This is achieved by minimizing the following entropy loss:
\begin{equation}
    \mathcal{L}_{e n t}\left(\mathbf{f} ; \mathbf{H}\right)  =-\frac{1}{N} \sum_{i=1}^N  \sum_{k=1}^C \delta_k\left(\mathbf{H}\left(f_i\right)\right) \log \delta_k\left(\mathbf{H}\left(f_i\right)\right)
\end{equation}
where $\delta_k\left(\mathbf{H}\left(f_i\right)\right)$ is the softmax score of class $k$ for the pseudo-source feature $f_i \in \mathbf{f}$. 

\begin{figure}[t]
    \centering
    \includegraphics[width=0.8\linewidth]{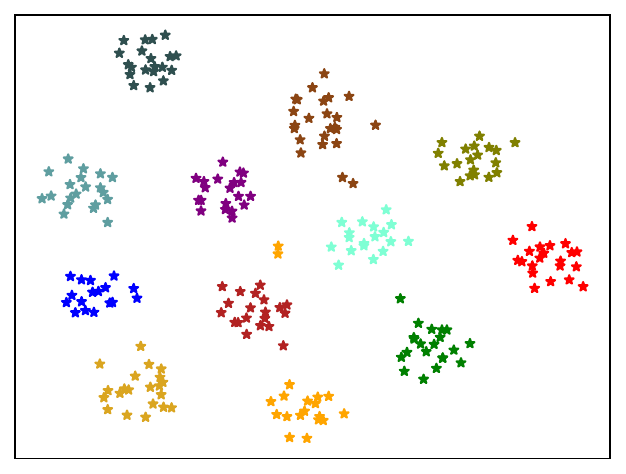}
    \caption{t-SNE plot of 240 generated pseudo-source features for TTA on VisDA dataset comprising of 12 classes.}
    \label{fig:gen_fea}
\end{figure}

Along with this, we use diversity maximization loss to avoid the trivial solution where all feature vectors collapse to the same class. 
% the feature vectors to be spread across all classes. In this way, we can ensure that each class is represented in the feature bank and 
This ensures there are adequate number of feature vectors from each class in $\mathbf{f}$. 
% This loss maximizes the distance between the feature vectors to achieve diversity.
\begin{equation}
\begin{aligned}
    \mathcal{L}_{d i v}\left(\mathbf{f} ; \mathbf{H}\right)  & =\sum_{k=1}^C \hat{p}_k \log \hat{p}_k \\
& =D_{K L}\left(\hat{p}, \frac{1}{C} \mathbf{1}_C\right)-\log C 
\end{aligned} 
\end{equation}
The loss is computed based on the mean softmax score of the test batch $\hat{p}=\mathbb{E}_{f \in \mathbf{f}}\left[\delta\left(h\left(f\right)\right)\right]$. 
The first term in the equation is the Kullback-Leibler (KL) divergence between the mean prediction vector $\hat{p}$ and the uniform distribution $\frac{1}{C} \mathbf{1}_C$. Here, $\hat{p}$ represents the marginal class distribution of the target data as estimated by the target model $\mathbf{F_t}$,  $C$ is the number of classes and $\mathbf{1}_C$ is a vector of ones with length $C$. The KL divergence measures the dissimilarity between two probability distributions, and in this context, it measures the discrepancy between the class distribution in the feature bank and the ideal case where all classes are equally represented.
Overall, the diversity maximization loss encourages the feature bank to have a balanced representation of features across all classes, which is important for improving the clustering performance of the TTA algorithm. To summarize, we optimize the following
\begin{equation}
\label{eq:fea_gen}
    \mathbf{f}^{*} = \argminA_{\mathbf{f}} \mathcal{L}_{ent}(\mathbf{f}; \mathbf{H}) + \beta \mathcal{L}_{div}(\mathbf{f}; \mathbf{H}) 
\end{equation}
% By optimizing the feature bank with both entropy and diversity losses, we gradually generate a feature bank with well-clustered and diverse samples. 
In Fig.(~\ref{fig:gen_fea}), we visualize the generated features on setting 20 samples per class for VisDA dataset.

\subsection{Pseudo Source Guided Target Clustering}
The use of feature bank has proven to be effective in Contrastive learning~\cite{moco} and SFDA methods like AaD~\cite{aad} and AdaContrast~\cite{adacontrast}. The proposed feature bank consists of pseudo-source features which are very different from the target feature bank used in ~\cite{aad, adacontrast}.
%The pseudo source feature bank however are well clustered. 
Unlike target features whose pseudo labels can be noisy, we can obtain clean labels for the generated pseudo-source features. 
We explain below how the generated features and their label information can be leveraged to better cluster and align the target features. We visually demonstrate the entire pSTarC framework in Fig.(~\ref{fig:protocols}).

\begin{table*}[t]
\centering
\footnotesize
\begin{adjustbox}{max width=\linewidth}
\begin{tabular}{@{}cccccccccccccc@{}}
\toprule
   Method       & plane & bycyl & bus & car & horse & knife & mcycl & person & plant & sktbrd & train & truck & Average  \\ \midrule
Source  & 57.2           & 11.1           & 42.4         & 66.9         & 55.0           & 4.4            & 81.1           & 27.3            & 57.9           & 29.4            & 86.7           & 5.8            & 43.8          \\
CAN$^{*}$ \cite{can} & 95.7 & 88.8 & 6.9 & 68.6 & 94.5 & 94.8 & 79.2 & 70.3 & 88.7 & 80.6 & 83.2 & 51.7 & 75.2 \\
MCC$^{*}$ \cite{mcc} & 93.9 & 78.4 & 70.4 & 74.3 & 92.5 & 84.2 & 84.5 & 58.2 & 86.6 & 36.0 & 86.1 & 20.6 & 72.2 \\
Source-Proxy TTA$^{*}$~\cite{src_proxy_icml23} & 92.5 & 82.4 & 85.8 & 74.2 & 92.7 & 88.5 & 83.9 & 85.8 & 92.8 & 62.5 & 75.2 & 32.5 & 79.1\\

BN-Adapt\cite{bn_neurips} & 87.3 & 52.1 & 83.7 & 52.8 & 83.7 & 57.0 & 83.6 & 59.2 & 69.1 & 54.7 & 80.0 & 28.1 & 66.0 \\
TENT\cite{tent} & 91.1 & 45.6 & 86.4 & 66.4 & 88.7 & 75.1 & 90.3 & 76.4 & 84.4 & 47.1 & 83.6 & 13.7 & 70.7 \\
AdaContrast~\cite{adacontrast}  & 95.0           & 68.0           & 82.7         & 69.6         & 94.3           & 80.8           & 90.3           & 79.6            & 90.6           & 69.7            & 87.6           & 36.0           & 78.7          \\

C-SFDA~\cite{csfda}   & 95.9   & 75.6  & 88.4  & 68.1   & 95.4 & 86.1    & 94.5  & 82.0  & 89.2  & 80.2 & 87.3 & 43.8 & {\bf 82.1}\\ 
\midrule
{\bf pSTarC}         & 95.1  & 82.1 & 83.6 & 61.2 & 93.8 & 89.9 & 87.9 & 80.7 & 90.9 & 81.9 & 87.6 & 48.1 & 81.9 \\ 
\bottomrule
\end{tabular}
\end{adjustbox}
\caption{Average class accuracy (\%) of pSTarC and other TTA methods on VisDA. $^*$ refers to methods utilizing source data to enable TTA.}
\label{tab: visda}
\end{table*}

Pseudo-labeling based on confidence thresholding has been used very effectively in several applications~\cite{fixmatch}.
Here, we propose a soft pseudo-labeling approach to cluster the target samples.
%, which we also demonstrate in Figure(~\ref{fig:protocols}). 
Specifically, we identify the low entropy test samples based on a threshold $\tau_t$, which we define as the mean entropy of the batch. We aim to align these selected test samples to the nearest pseudo-source samples which belong to the same class as the sample. 
%Thresolding based pseudo-labeling has been used very effectively in semi-supervised learning methods like FixMatch~\cite{fixmatch}, where they use pseudo label loss for confident unlabeled data along with cross entropy loss for labeled data. 
%In our case, we do not have any labeled data and hard pseudo labeling the test samples is undesirable. Instead, we propose a soft pseudo labeling method to cluster the target samples. We identify the confident test samples based on a threshold $\tau$ and aim to align them to the nearest pseudo source samples which belong to the same class as the sample's prediction. 
%We now formally describe the pseudo-source guided clustering process.
Formally, given the generated feature bank $\mathbf{f}^{*}$, we first obtain their softmax score vectors and pseudo labels.
We denote $p_{i} = \delta(\mathbf{H}(f_{i})) $ as the softmax score vector and $\hat{y}_{i} = \argmaxA _{c} p_{i,c}$ as the pseudo label for feature $f_i$, where $p_{i,c}$ is the score of feature $i$ for class $c$. We partition the features into sets $S_{c}$ based on their pseudo labels as follows:
\begin{equation}
    S_{c} = \{f_{i}; \quad \hat{y}_{i} = c, f_{i} \in \mathbf{f}\}; \quad c \in \{1 \ldots C\}
\end{equation}
These sets are obtained once for the pseudo-source features generated and kept fixed throughout the adaptation process.

Given a test batch $\mathbf{x}_t$, we first obtain their confidence scores and pseudo labels and set the threshold $\tau_{t} = \mathbb{E}_{x_k \in \mathbf{x}_{t}}[e_k]$, the mean entropy of the batch. 
For a test sample $x_k \in \mathbf{x}_{t}$ (test batch), we denote its pseudo label as $\hat{y}_{k}$ and compute the sample entropy as $e_k$. %$s_k = \sum_{c} p_{k,c} \log p_{k,c}$. 
% Formally, for each test sample $x_t$ whose prediction by the model is $\hat{y}$, 
For this sample, we define its positive set $\mathbf{p}^{+}$ based on its entropy $e_k$ as follows: (1) When $e_{k}<\tau_{t}$, we define the positives to be $K$ nearest pseudo-source samples from set $S_{\hat{y}_{k}}$. 
(2) For samples which have high entropy, i.e, with $e_{k}>\tau_t$, as the pseudo labels can be highly noisy, it is not desirable to enforce them to align towards any pseudo-source samples. Instead, we anchor it to its own prediction vector by setting $\mathbf{p}^{+}=\{p_{k}\}$. 
In addition, we use its strong image augmentation $\tilde{x}_k$ to enforce prediction consistency between $p_k$ and $\tilde{p}_{k}$, the prediction vector of $\tilde{x}_k$. This helps the model to be invariant to image transformations and improves its generalization ability. We also use the dispersion loss that makes a sample dissimilar to the other samples in the batch, which is representative of the test data in all. This dispersion loss prevents the model from the trivial solution of all test samples collapsing to the same class. 
Our objective now is to make the predictions of the target embeddings similar to its positives without facing mode collapse, which we achieve by optimizing the following loss:
\begin{equation}
\label{eq:pstar}
    % \mathcal{L}_{\textrm{pSTarC}}(x_k) = -p_{k}^{T}\tilde{p}_{k} - \sum_{p_{j}^{+}\in \mathbf{p}^{+}} p_{k}^{T}p_{j}^{+} + \alpha \sum_{x_{j}\in\mathbf{x}_t} p_{k}^{T}p_{j} 
\mathcal{L}_{\textrm{pSTarC}}(x_k) =\underbrace{\vphantom{\sum_{p_{j}^{+}\in \mathbf{p}^{+}} p_{k}^{T}p_{j}^{+}}-p_{k}^{T}\tilde{p}_{k}}_\text{$L_{aug}$} - \underbrace{\sum_{p_{j}^{+}\in \mathbf{p}^{+}} p_{k}^{T}p_{j}^{+}}_{\text{$L_{attr}$}} + \underbrace{\vphantom{\sum_{p_{j}^{+}\in \mathbf{p}^{+}} p_{k}^{T}p_{j}^{+}}\lambda \sum_{x_{j}\in\mathbf{x}_t} p_{k}^{T}p_{j} }_\text{$L_{disp}$} 
\end{equation}
We perform one step optimization on test batch $\mathbf{x}_t$ using this loss and then predict their labels. This process is repeated for each batch in the TTA setting. \\

\begin{table*}[t]
\footnotesize
\centering
\begin{adjustbox}{max width=\linewidth}
\label{tab: office-home}
\begin{tabular}{@{}cccccccccccccc@{}}
\toprule
Method     & A $\rightarrow$ C & A $\rightarrow$ P & A $\rightarrow$ R & C $\rightarrow$ A & C $\rightarrow$ P & C $\rightarrow$ R & P $\rightarrow$ A & P $\rightarrow$ C & P $\rightarrow$ R & R $\rightarrow$ A & R $\rightarrow$ C & R $\rightarrow$ P & Average \\ \midrule
Source                             & 44.6                                & 66.5                                 & 73.5                                & 51.0                                & 61.9                                & 63.2                                & 51.1                                & 40.5                                & 71.9                                & 64.4                                 & 47.1                                & 77.3                                & 59.4\\
BN-Adapt~\cite{bn_neurips}  & 38.9 &	59.9 &	71.5 &	55.0 &	62.0 &	65.2 &	54.4 &	37.3 &	71.6 &	65.2 &	41.3 &	73.8 &	58.0\\
TENT~\cite{tent} & 39.1 &	60.2 &	71.6 &	55.2 &	62.2 &	65.5 &	54.6 &	37.6 &	71.8 &	65.3 &	41.6 &	73.9 &	58.2\\
AdaContrast~\cite{adacontrast} & 42.2 &	64.5 &	73.2 &	56.2 &	64.1 &	66.4 &	54.7 &	40.4 &	73.0 &	66.7 &	45.1 &	75.6 &	60.2  \\
\midrule
{\bf pSTarC} & 47.7 &	68.7 &	75.4 &	58.6 &	68.4 &	68.9 &	55.1 &	45.8 &	75.6 &	67.5 &	51.8 &	78.7 & \textbf{63.5} \\ \bottomrule 

\end{tabular}
\end{adjustbox}
\caption{Total accuracy (\%) of pSTarC and other TTA methods on Office-Home dataset.}
\label{tab: office-home}
\end{table*}

\begin{table*}[t!]
\begin{adjustbox}{max width=\linewidth}
\begin{tabular}{ccccccccccccccccc}
\toprule
Method       & gaussian & shot & impulse & defocus & glass & motion & zoom & snow & frost & fog  & brightness & contrast & elastic & pixelate & jpeg & Average         \\ \midrule
Source &	27 &	32 &	60.6 &	70.7 &	45.9 &	69.2 &	71.2 &	60.5 &	54.2 &	49.7 &	70.5 &	44.9 &	62.8 &	25.3 &	58.8 &	53.6 \\
BN Adapt~\cite{bn_neurips}	& 57.9 &	59.3 &	57.3 &	 72.4 &	58.1 &	70.3 &	72.1 &	65.1 &	65 &	58.5 &	73.5 &	69.7 &	64.3 &	67.1 &	58.8 &	64.6 \\
TENT~\cite{tent} & 62.7	& 65.1 & 65.5 &	75.0 &	62.6 &	72.5 &	75.0 &	69.6 &	68.1 &	66.2 &	76.0 &	71.8 &	67.1 &	71.6 &	63.1 &	68.8	 \\
AdaContrast~\cite{adacontrast} & 57.3 &	59.4 &	61.1 &	73.4 &	58.8 &	71.1 &	73.4 &	66.6 &	67.3 &	60.7 &	75.2 &	71.8 &	65.4 &	65.8 &	60.5 &	65.9 \\
\midrule
{\bf pSTarC} & 63.4 &	65.4 &	66.5 &	75 &	63 & 73.2 &	74.9 &	70.3 &	69.8 &	66.5 &	76.6 &	73.2 &	68.0 &	72.2 &	63.8 &	\textbf{69.5} \\
\bottomrule
\end{tabular}
\end{adjustbox}
\caption{Accuracy (\%) of different TTA methods on 15 corruptions from CIFAR-100C dataset in TTA setting.}
\label{tab:CIFAR-100C-tta}
\end{table*}

\noindent\textbf{What makes pSTarC an effective framework?} \\
\indent 1. We operate in the {\em fully test-time scenario}, i.e., we do not assume access to source data in any form unlike some prior methods~\cite{src_proxy_icml23, can, mcc}, which use the source data to equip the model for future TTA.  
    %This may not be always feasible in a practical scenario due to privacy issues. 
    In pSTarC, we leverage the classifier which is a part of the given source model to synthesize pseudo-source features to enable clustering during test time. \\
\indent 2. Feature banks have been effectively used in AdaContrast\cite{adacontrast} to cluster the test data. However, it is expensive to have multiple large memory buffers which have to be continuously updated. We propose a {\em simple one-step pseudo source generation} framework. These generated features can be used during TTA forever, as the final goal indeed is to align the test distribution to the source distribution. \\
% so that the classifier can then be effectively translated to evaluate the test data.
\indent 3. pSTarC is a {\em memory efficient framework} as we only store the online updating model, in contrast to AdaContrast\cite{adacontrast} and C-SFDA\cite{csfda} where they need to store the student and teacher model. Our framework is also {\em more efficient in runtime} as we only forward pass the image and its strong augmentation, while the state-of-the-art method C-SFDA~\cite{csfda} uses 12 augmentations.

\begin{table}[t!]
\begin{adjustbox}{max width=\linewidth}
% \footnotesize
\centering
\begin{tabular}{ccccccccc}
\toprule
Method      & R$\rightarrow$C  & R$\rightarrow$P  & P$\rightarrow$C  & C$\rightarrow$S  & S$\rightarrow$P  & R$\rightarrow$S  & P$\rightarrow$R  & Average \\ \midrule
Source      & 55.5 & 62.7 & 53   & 46.9 & 47.3 & 46.3 & 75.0   & 55.2    \\
BN-Adapt~\cite{bn_neurips} & 54.1 &	62.8 &	54.3 &	49.4 &	59.1 &	47.6 &	75.0 &	57.5     \\
% BN-Adapt(S+T) &	54.8 &	63.2 &	54.2 &	50.1 & 60.6	& 48.3 & 75.6 &	57.7 \\
TENT~\cite{tent}        &  55.6 &	64.5 &	55.5 &	50.8 &	59.9 &	49.9 &	75.9 &	58.9   \\
% BN-Adapt (S+T)     & 54.6 & 63.4 & 54.5 & 49.5 & 59.4 & 48.1 & 75.4 & 57.8    \\
% TENT (S+T)   & 57.1 & 65.5 & 56.3 & 51.4 & 60.4 & 51.5 & 76.2 & 59.8    \\
% AaD-TTA     & 55.9 & 64   & 55.5 & 51.1 & 61.1 & 49.4 & 76.9 & 59.1    \\
AdaContrast~\cite{adacontrast} & 61.1 & 66.9 & 60.8 & 53.4 & 62.7 & 54.5 & 78.9 & 62.6    \\ 
C-SFDA~\cite{csfda} & 61.6 &	67.4 &	61.3 &	55.1 &	63.2 &	54.8 &	78.5 &	63.1 \\ \midrule
{\bf pSTarC} & 60.8 &	67.7 &	60.3 &	55.6 &	65.3 &	55.8 &	80.2 &	\textbf{63.7} \\ \bottomrule
\end{tabular}
\end{adjustbox}
\caption{Total accuracy (\%) of TTA methods on DomainNet-126.}
\label{tab:domainnet}
\end{table}

\begin{table*}[t!]
\begin{adjustbox}{max width=\linewidth}
\begin{tabular}{ccccccccccccccccc}
\toprule
Method       & gaussian & shot & impulse & defocus & glass & motion & zoom & snow & frost & fog  & brightness & contrast & elastic & pixelate & jpeg & Average         \\ \midrule
Source &	27 &	32 &	60.6 &	70.7 &	45.9 &	69.2 &	71.2 &	60.5 &	54.2 &	49.7 &	70.5 &	44.9 &	62.8 &	25.3 &	58.8 &	53.6 \\
BN Adapt~\cite{bn_neurips}	& 57.9 &	59.3 &	57.3 &	 72.4 &	58.1 &	70.3 &	72.1 &	65.1 &	65 &	58.5 &	73.5 &	69.7 &	64.3 &	67.1 &	58.8 &	64.6 \\
TENT~\cite{tent} &	62.8 &	64.2 &	58.3 &	62.1 &	48.8 &	51.7 &	51.5 &	41.6 &	36.3 &	28.9 &	29.6 &	17.7 &	12.0 & 	11.5 &	9.6 &	39.1 \\
CoTTA~\cite{cotta} &	59.9 &	62.3 &	60.3 &	73.1 &	62.0 &	72.1 &	73.6 &	67.2 &	68.2 &	59.7 &	75.3 &	73.1 &	67.5 &	71.7 &	66.5 &	67.5 \\
AdaContrast~\cite{adacontrast} & 57.7 &	63.2 &	61.4 &	72.3 &	59.9 &	70.9 &	72.5 &	67.1 &	69.3 &	61.8 &	74.1 &	71.7 &	66.1 &	66.7 &	63.8 &	66.6 \\
RMT~\cite{rmt}	& 59.5 &	63.9 &	63.7 &	72.3 &	66.1 &	71.5 &	73.6 &	71.0 &	71.0 &	67.5 &	74.9 &	72.6 &	71.8 &	73.7 &	70.7 &	\textbf{69.6} \\
% PETAL	61.7	63.6	61.4	74.1	63.3	72.8	74.6	68	69.2	61.3	75.6	73.58	68.5	73.1	67.5	68.54
% SANTA	63.5	66.9	64.9	74.1	65.1	72.3	74.6	70.5	70.1	66.9	76.4	73.3	68.1	72.5	64.8	69.7
\midrule
{\bf pSTarC} & 63.4 &	67.0 &	64.0 &	71.1 &	62.9 &	69.3 &	72.4 &	67.3 &	68.7 &	64.1 &	72.9 &	71.9 &	66.7 &	70.5 &	62.9 &	67.7 \\
\bottomrule
\end{tabular}
\end{adjustbox}
\caption{Accuracy (\%) of different methods on 15 corruptions from CIFAR-100C dataset in CTTA setting.}
\label{tab:CIFAR-100C}
\end{table*}

\section{Experimental Evaluation}
\label{expt}
We evaluate the proposed framework extensively on three real-world domain shift datasets, namely VisDA ~\cite{visda}, 
DomainNet-126~\cite{moment} and Office-Home~\cite{office-home} and also on a corruption benchmark dataset, namely CIFAR100C~\cite{cifar-100c}. \\

\noindent{\bf Datasets:} 
\textbf{VisDA} is a challenging dataset for object recognition tasks with synthetic to real domain shift. The target domain consists of $55,388$ real object images from $12$ classes.
\textbf{Office-Home} contains four domains - Real, Clipart, Art, Product and $65$ classes with a total of $15,500$ images.
\textbf{DomainNet-126} is a subset of DomainNet consisting of $126$ classes from four domains, namely Real, Sketch, Clipart and Painting.
\textbf{CIFAR-100C} is a corruption benchmark with domain shifts like gaussian noise, blur, weather changes, etc.  
% Each corruption has five levels of severity, applied to the test set of CIFAR100. 
Following~\cite{cotta}, we use severity level 5 corruptions. 
For VisDA-C, we compare the average of per-class accuracies while for the other datasets, we compare the average of total accuracy across domain shifts. \\

\noindent{\bf Model Architecture:}
For TTA experiments, we use ResNet-50\cite{resnet} as the backbone for Office-Home and DomainNet-126 datasets and ResNet-101\cite{resnet} for the VisDA dataset. We use the same network architecture as in~\cite{adacontrast}, in which the final part of the network is modified to include fully connected layer and Batch Normalization, and then followed by a classifier, which is a fully connected layer with weight normalization. 
For CIFAR-100C , we use ResNeXt\cite{resnext} as used in~\cite{cotta, rmt}. \\

\noindent\textbf{Implementation details:}
We use Pytorch framework and run all experiments on a single NVIDIA A-5000 GPU. 
% We use the source model trained using the same protocol as described in ~\cite{adacontrast, csfda, cotta} for fair comparison.
For source training, following~\cite{adacontrast, csfda} the model is initialized with ImageNet pre-trained weights and trained for $10$, $60$ and $50$ epochs for VisDA, DomainNet-126 and Office-Home respectively. 
During test time adaptation, we only update the backbone parameters, keeping the classifier fixed for all experiments. 
Following~\cite{adacontrast, src_proxy_icml23, csfda}, we set the batch size to $128$ in all experiments for VisDA, DomainNet-126 and Office-Home. We use SGD as the optimizer with learning rate of 5e-4 and momentum $0.9$. 
Following \cite{cotta, rmt}, for CIFAR-100C, the batch size is set to $200$ and we use Adam~\cite{adam} optimizer with learning rate of 1e-3.
We set $\beta$ to $5$ in eqn.(\ref{eq:fea_gen}) and the number of features per class $n_c$ to 20 in all experiments.
We report the results of prior methods from the respective papers. 
We use the official code provided by AdaContrast~\cite{adacontrast} to perform experiments on Office-Home and also adapt it to CTTA setting.
In the Supplementary material, we describe the image augmentations used, analysis on parameter $n_c$ and provide the pseudo code for pSTarC. 
\color{black}

\begin{table}[t]
\centering
\footnotesize
\begin{adjustbox}{max width=\linewidth}
    \begin{tabular}{@{}cccccc@{}}
    \toprule
    Method         & {\rotatebox{60}{Real$\rightarrow$}} & {\rotatebox{60}{Clipart$\rightarrow$}} & {\rotatebox{60}{Painting$\rightarrow$}} & {\rotatebox{60}{Sketch$\rightarrow$}} & Average \\ \midrule
    Source only    & 54.7 & 50.7 & 58.3 & 55.2 & 54.7   \\
    BN Adapt~\cite{bn_neurips}       & 54.9 & 54.8 & 60.5 & 62.2 & 58.1   \\
    TENT~\cite{tent} & 57.6 & 55.8 & 62.8 & 62.5 & 59.7   \\
    CoTTA~\cite{cotta}          & 56.6 & 57.0 & 63.6 & 63.7 & 60.2   \\
    AdaContrast~\cite{adacontrast}    & 62.2 & 62.4 & 67.7 & 68.1 & 65.1   \\
    RMT~\cite{rmt}            & 63.0 & 62.1 & 68.3 & 67.9 & 65.3   \\ \midrule
    {\bf pSTarC}         & 62.7 & 63.6 & 67.6 & 68.1 & \textbf{65.5}   \\ \bottomrule
    \end{tabular}
\end{adjustbox}
\caption{Accuracy (\%) of different TTA methods on four domain shift sequences from DomainNet-126 in CTTA setting.}
\label{tab:domainnet_ctta}
\end{table}
%\vspace{-0.2 cm}

% ************************
\subsection{Evaluation for TTA setting}
We compare the performance of our proposed pSTarC framework with the prior TTA approaches~\cite{bn_neurips, tent, adacontrast, csfda, src_proxy_icml23}. 
For VisDA dataset, from Table~\ref{tab: visda}, we observe that pSTarC performs at par with the state-of-the-art method C-SFDA~\cite{csfda}, while being computationally much more efficient (Table~\ref{tab:complexity}). 
Interestingly, it also outperforms the approaches which assume access to the source data before performing TTA.
On Office-Home, we get a significant improvement of 3.5\% compared to the prior TTA method AdaContrast~\cite{adacontrast} as shown in Table~\ref{tab: office-home}. 
% C-SFDA~\cite{csfda} does not report results on TTA setting for Office-Home and hence we do not compare with it. 
On DomainNet-126, from Table~\ref{tab:domainnet}, we observe that pSTarC achieves an average accuracy of 63.7\% across 7 domain shifts, outperforming all the existing approaches including~\cite{csfda}.
On CIFAR-100C~\cite{cifar-100c}, our method performs 1.1\% better than TENT~\cite{tent} and 3.6\% better than AdaContrast~\cite{adacontrast}, suggesting its effectiveness even on corruption domain shifts (Table~\ref{tab:CIFAR-100C-tta}).
% ************************
\subsection{Evaluation for CTTA setting}
%\noindent{\bf Performance of pSTarC in CTTA:}
We also study the effectiveness of pSTarC in the CTTA setting where test domains change with time. 
% Following CoTTA~\cite{cotta}, for CTTA on CIFAR-100C, the corruptions are changed from gaussian noise to jpeg compression as shown in Table~\ref{tab:CIFAR-100C}. We outperform AdaContrast~\cite{adacontrast} by 1.1\% and the SOTA CTTA method CoTTA~\cite{cotta}. In addition to the evaluation on the corruption benchmark, 
% we also perform experiments on continual domain shifts synthesized from DomainNet-126. 
To do this, we perform experiments on CIFAR-100C and the following four domain sequences from DomainNet-126: \\
(1) {\em Real}-World$\rightarrow$Clipart$\rightarrow$Painting$\rightarrow$Sketch; \\
(2) {\em Clipart}$\rightarrow$Sketch$\rightarrow$Real-World$\rightarrow$Painting; \\ 
(3) {\em Painting}$\rightarrow$Real-World$\rightarrow$Sketch$\rightarrow$Clipart \\
(4) {\em Sketch}$\rightarrow$Painting$\rightarrow$Clipart$\rightarrow$Real-World. 

The {\em first domain} indicates the source domain, which is then adapted to the other three test domains in the above sequence.
From Table ~\ref{tab:domainnet_ctta}, we observe that pSTarC outperforms all the state-of-the-art approaches in this challenging setting. 
Specifically, it outperforms CoTTA by a significant margin of 5.3\% and also performs favourably compared to the state-of-the-art method RMT~\cite{rmt}.
In addition, we also evaluate pSTarC on CIFAR-100C continual setting and report the results in Table~\ref{tab:CIFAR-100C}. 
It performs favourably compared to AdaContrast~\cite{adacontrast} and CoTTA~\cite{cotta}, while RMT~\cite{rmt} performs the best in this case.
% In the Supplementary Material, we also show that the proposed method works satisfactorily for the corruption benchmark dataset, namely CIFAR-100C~\cite{cifar-100c} for the CTTA task.
% In summary, these experiments  demonstrate the consistent effectiveness of our method across diverse domain shifts in both single domain and continual test time adaptation settings.   
But, CoTTA~\cite{cotta} and RMT~\cite{rmt} are computationally more expensive as they need to store teacher and student models, while pSTarC is more light-weight as it only stores one model.

In Figure~\ref{fig:spiderweb}, we summarize the performance of pSTarC with the source model, TENT~\cite{tent} and AdaContrast~\cite{adacontrast}. In this plot, the lines farther from the center indicates better performance. We observe that pSTarC outperforms these methods across all domain shifts for both TTA and CTTA.

\begin{table}[t]
\centering
\footnotesize
\begin{tabular}{@{}ccccc@{}}
\toprule
$L_{aug}$  & $L_{attr}$ & $L_{disp}$ & VisDA & DomainNet-126 \\ \midrule
  \cmark   & \cmark      &            & 68.8            &  58.8         \\
  \cmark   &             & \cmark     & 78.2            &  59.7         \\ 
           &  \cmark     & \cmark     & 80.0            &  63.0         \\ 
  \cmark   & \cmark      & \cmark     & \textbf{81.9}   &  \textbf{63.7}  \\ \bottomrule
\end{tabular}
\caption{Ablation study: Importance of each loss term.}
\label{tab:loss_ablation}
\end{table}

% ******************************************
\subsection{Additional Analysis}
Here, we report the results of additional analysis to better understand the proposed framework. \\ 
{\bf Ablation Study:}
The proposed pSTarC framework consists of three loss components. The first component is $L_{aug}$ which enforces consistency between an image and its augmentation. From Table~\ref{tab:loss_ablation}, we observe that using strong augmentations can indeed help improve the feature representations, as we get 1.9\% and 0.7\% improvement on VisDA and DomainNet-126 respectively. 
The second component $L_{attr}$ aims to align the test features with the pseudo source features. On removing the attraction loss component from $L_{pSTarC}$, the loss becomes similar to contrastive learning. While this performs reasonably, achieving 78.2\% and 59.7\% on VisDA and DomainNet respectively, incorporating the pseudo-source features improves the results significantly by 3.7\% and 4\%,  proving that they indeed help model adaptation by correctly aligning the test features so that the source trained classifier can well classify the test data.    The third component, $L_{disp}$ is the dispersion term which prevents the model to avoid all the test features collapsing to one cluster, which is a trivial solution when optimizing only the attraction loss $L_{attr}$. This term plays a role similar to the diversity term and is crucial in any unsupervised adaptation protocols~\cite{shot, aad} to avoid model collapse, the effect of which we observe in Table~\ref{tab:loss_ablation}. The accuracy on VisDA and DomainNet-126 drop to 68.8\% and 58.8\% respectively, as the test samples would be predicted into lesser number of classes than actually present in the dataset. \\

\noindent{\bf Performance on varying batch sizes:}
In TTA, it is crucial for the method to be able to continuously adapt even with very few samples. 
In this analysis, we vary the batch size from $8$ to $128$ and perform experiments on the DomainNet-126 dataset. 
Table\ref{tab:batch_size} reports the average accuracy across $7$ domain shifts for each batch size. 
We observe that the proposed pSTarC consistently outperforms both TENT~\cite{tent} and AdaContrast~\cite{adacontrast} for all batch sizes. 
The effect is more pronounced for the smallest batch size $8$, where pSTarC outperforms TENT by a huge margin of 15.3\% and AdaContrast by 4\%. On average, pSTarC does better than TENT by 6.2\% and AdaContrast by 1.7\%. \\

\input{spiderweb}

\noindent {\bf Complexity Analysis:}
Here, we analyse the complexity of pSTarC and three other recent TTA methods:  AdaContrast~\cite{adacontrast},  Source-Proxy-TTA~\cite{src_proxy_icml23}~ and C-SFDA~\cite{csfda} on VisDA dataset.
In the TTA setting, it is desirable to have methods that requires storing less additional information due to memory limitations and privacy concerns.
The prior methods AdaContrast~\cite{adacontrast} and C-SFDA~\cite{csfda} are based on the teacher student framework. Hence, it needs to store twice the number of model parameters, while we only store the updating model parameters in pSTarC, as we report in Table~\ref{tab:complexity}. 
AdaContrast stores a memory queue of size $16384$ to collect key features (of dimension 256), and its pseudo labels, which is used to retrieve positives for contrastive learning. Alongside, they store another feature bank (of size 1024) and their corresponding scores which is used to retrieve neighbours for soft pseudo-labeling the target samples. 
Thus, the total memory buffer required for AdaContrast is 16384x(256+1)+1024x(256+12). 
\cite{src_proxy_icml23} condenses the source data to save 25 images per class of size 112x112 for VisDA dataset. This accounts to a memory requirement of 37.6M (12x25x112x112).
On the other hand, in the pSTarC framework, we only store $20$ features per class and the corresponding scores resulting in a memory buffer of $240\text{x}(256+12)$. C-SFDA does not store any features or images. 
However, they need $13$ forward passes (12 augmentations in addition to the actual test sample), while  AdaContrast~\cite{adacontrast} and Source-Proxy-TTA~\cite{src_proxy_icml23} uses 3 augmentations, and pSTarC uses only two augmentations.
We summarize this in Table~\ref{tab:complexity}, which shows that pSTarC is very efficient, in addition to achieving better or performance comparable to the state-of-the-art across several challenging settings.

\begin{table}[]
\centering
\footnotesize
\begin{tabular}{@{}ccrrrrc@{}}
\toprule
\multirow{2}{*}{Method} & \multicolumn{5}{c}{Batch size}                                                                                                & \multirow{2}{*}{Average}          \\
                        & 8                        & \multicolumn{1}{c}{16} & \multicolumn{1}{c}{32} & \multicolumn{1}{c}{64} & \multicolumn{1}{c}{128} &                                   \\ \midrule
TENT                    & 38.8 & 55.4 & 58.6 & 59.1 & 58.9 & \multicolumn{1}{r}{54.2} \\
AdaContrast             & \multicolumn{1}{r}{50.1} & 57.9                   & 60.8                   & 62.4                   & 62.4                    & \multicolumn{1}{r}{58.7}          \\ \midrule
{\bf pSTarC}                  & \multicolumn{1}{r}{54.1} & 59.2                   & 61.3                   & 63.8                   & 63.7                    & \multicolumn{1}{r}{\textbf{60.4}} \\ \bottomrule
\end{tabular}
\caption{Ablation on batch size using DomainNet-126}
\label{tab:batch_size}
\end{table}

\begin{table}[t!]
\centering
\begin{adjustbox}{max width = \linewidth}
\begin{tabular}{@{}ccccc@{}}
\toprule
Method       & AdaContrast & Source-Proxy-TTA & C-SFDA & pSTarC     \\
\midrule
% \#Parameters & 86,050,432 & 43,025,216  & 43,025,216 \\
\#Parameters & 86M & 43M & 86M & 43M \\
Memory    & 4.67M  & 3.76M & -   & 0.03M      \\
% Memory    & 4665344  & 3763200 & -   & 32160      \\
\#Forward    & 3     & 3   & 13  & 2          \\
\#Backward   & 1     & 1   & 1  & 1          \\
\bottomrule
\end{tabular}
\end{adjustbox}
\caption{Complexity Analysis of TTA methods on VisDA}
\label{tab:complexity}
\end{table}

\section{Conclusion}
In this paper, we have proposed a novel framework termed pSTarC for Test Time Adaptation (TTA) of deep neural networks.
pSTarC leverages the fixed source classifier to generate pseudo-source samples, which is then used to align the test samples, which enables the source trained classifier to classify test data from different distributions. 
Extensive experiments on several real-world domain shift datasets justify the effectiveness of our proposed framework. %We outperform the state-of-the-art datasets on most settings. 
Additionally, we also show that the method can seamlessly be used in continual test time adaptation scenario, though there is still scope for improvement in the corruption datasets. 
%Although, we acknowledge that there is scope for improvement in this sceanario, when comapred to methods specifically designed for CTTA setting.
Overall, our findings highlight the importance of target clustering techniques and leveraging the source classifier for improving test-time adaptation performance in several real-world challenging scenarios. \\

\noindent \textbf{Acknowledgements} 
This work is partly supported through a research grant from SERB (SPF/2021/000118), Govt. of India. The first author is supported by Prime Minister's Research Fellowship awarded by Govt. of India.

%%%%%%%%% REFERENCES
{\small
\bibliographystyle{ieee_fullname}
\bibliography{arXiv_Submission}
}

\end{document}

\title{Supplementary Material} 

\maketitle

\section{Analysis on DomainNet-126}
Here, we provide detailed analysis done on DomainNet-126. The paper results the average accuracy across the seven domains in Table 7 and 8 in the main paper. Here, we provide the results for each domain shift on performing ablation on loss components in Table~\ref{tab:loss_ablation}. We also report detailed results of AdaContrast (Table~\ref{tab:adacon_batch}) and pSTarC (Table~\ref{tab:pstarc_batch}) on varying batch sizes.

\setlength{\tabcolsep}{3pt}
\begin{table}[h]
\centering
\footnotesize
\begin{adjustbox}{max width=\linewidth}
\begin{tabular}{@{}ccccccccccc@{}}
\toprule
$L_{aug}$  & $L_{attr}$ & $L_{disp}$ & R$\rightarrow$C  & R$\rightarrow$P  & P$\rightarrow$C  & C$\rightarrow$S  & S$\rightarrow$P  & R$\rightarrow$S  & P$\rightarrow$R  & Avg \\ \midrule
  \cmark   & \cmark      &            & 57.1 &	66.6 &	57.1 &	47.8 &	54.7 &	54.1 &	74.0 &	58.8  \\
  \cmark   &             & \cmark     & 56.0 &	63.6 &	56.0 &	51.8 &	61.7 &	49.9 &	78.7 &	59.7      \\ 
           &  \cmark     & \cmark     & 60.1 &	67.3 &	59.6 &	55.0 &	64.8 &	54.6 &	79.8 &	63.0\\ 
  \cmark   & \cmark      & \cmark     & \textbf{60.8} &	\textbf{67.7} &	\textbf{60.3} &	\textbf{55.6} &	\textbf{65.3} &	\textbf{55.8} &	\textbf{80.2} &	\textbf{63.7}  \\ \bottomrule
\end{tabular}
\end{adjustbox}
\caption{pSTarC ablation study: Importance of each loss term.}
\label{tab:loss_ablation}
\end{table}

\setlength{\tabcolsep}{6pt}
\begin{table}[h]
\begin{adjustbox}{max width=\linewidth}
% \footnotesize
\centering
\begin{tabular}{ccccccccc}
\toprule
Method      & R$\rightarrow$C  & R$\rightarrow$P  & P$\rightarrow$C  & C$\rightarrow$S  & S$\rightarrow$P  & R$\rightarrow$S  & P$\rightarrow$R  & Average \\ \midrule
8  & 44.9 &   54.7 &	45.8 &	44.1 &	52.1 &	42.2 &	66.9 &	50.1 \\
16 & 54.4 &	62.8 &	54.6 &	49.6 &	59.3 &	49.3 &	75.2 &	57.9\\
32 & 57.8 & 65.4 &	58.1 &	52.0 &	61.9 &	52.9 &	77.6 &	60.8\\
64 & 60.0 &	66.9 &	59.7 &	53.6 &	63.7 &	54.3 &	78.6 &	62.4\\
128 & 60.1 &	67.1 &	60.2 &	54.4 &	64.2 &	54.3 &	78.7 &	62.4 \\
% 128 & 60.8 & 67.7 &	60.3 &	55.6 &	65.3 &	55.8 &	80.2 &	63.7\\
\bottomrule
\end{tabular}
\end{adjustbox}
\caption{Total accuracy (\%) of AdaContrast on varying batch sizes.}
\label{tab:adacon_batch}
\end{table}

\begin{table}[h]
\begin{adjustbox}{max width=\linewidth}
% \footnotesize
\centering
\begin{tabular}{ccccccccc}
\toprule
Method      & R$\rightarrow$C  & R$\rightarrow$P  & P$\rightarrow$C  & C$\rightarrow$S  & S$\rightarrow$P  & R$\rightarrow$S  & P$\rightarrow$R  & Average \\ \midrule
8  & 53.5 &	62.6 &	51.2 &	41.2 &	54.1 &	46.2 &	69.9 &	54.1\\
16 & 56.5 &	65.7 &	56.2 &	49.6 &	59.8 &	51.4 &	75.5 &	59.2\\
32 & 59.4 &	67.2 &	58.0 &	51.1 &	61.9 &	54.5 &	77.0 &	61.3\\
64 & 61.6 &	68.9 &	60.5 &	54.7 &	64.3 &	57.0 &	79.4 &	63.8\\
128 & 60.8 & 67.7 &	60.3 &	55.6 &	65.3 &	55.8 &	80.2 &	63.7\\
\bottomrule
\end{tabular}
\end{adjustbox}
\caption{Total accuracy (\%) of pSTarC on varying batch sizes.}
\label{tab:pstarc_batch}
\end{table}

\section{Choice of parameter $n_c$}

For pseudo source feature generation, we set the total number of features $N$ as $C\times n_c$ where $C$ is the number of classes and $n_c$ is the number of features we expect to be generated per class. 
During test time adaptation, we set the number of positives $K$ as 5 for all experiments. So, ensuring the generated feature bank to contain 5 samples per class should suffice for the algorithm to work well without any significant degradation in accuracy. 

We use an Adam optimizer with a learning rate of 0.01 and optimize the feature bank for 50 steps.
However, as we are optimising the feature bank, naively setting $N=5 \times C$ may not ensure there are adequate number of features (5 in this case) per class. 
The optimum for the second term $L_{div}$ in the loss occurs only when there are equal number of samples in each class, i.e., when the class distribution become uniform. This can not always be guaranteed, while using the same optimization parameters across datasets. For example, even setting 120 as the number of features for VisDA with 12 classes, atleast 5 features per class were generated. But for DomainNet-126, using the same optimization scheme, we observed some classes had less than 5 features generated. Instead of tuning the optimizer hyperparameters for each dataset, we set $n_c$ sufficiently large (20 here) so that it can be used across all datasets.  We observe that on setting $N$ as $20\times C$ features, for all datasets, using the same optimizer parameters and number of steps, we could ensure atleast 5 features per class were generated to enable using them seamlessly during TTA. Hence, we set $N$ to 20 features per class for all datasets, VisDA (with 12 classes), DomainNet-126 (with 126 classes), Office-Home (with 65 classes) and CIFAR-100 (with 100 classes).

\section{Pseudo Code for pSTarC}

\begin{lstlisting}[language=Python, numbers=none]

def generate_features(args, netC, num_features=100, num_epochs=50, feature_dim=256):
    netC.train()
    pseudo_features = torch.randn((args.class_num * num_features, feature_dim)).cuda()
    pseudo_features.requires_grad = True
    optim_feats  = optim.Adam ([pseudo_features], lr =0.01)
    for t in range(num_epochs):
        optim_feats.zero_grad()
        scores = nn.Softmax(dim=1)(netC(pseudo_features))
        loss_ent = torch.mean(Entropy(scores))        
        loss_div = +torch.sum(torch.mean(scores, 0) * torch.log(torch.mean(scores, 0) + 1e-6))
        loss = loss_ent + loss_div *5        
        loss.backward()
        optim_feats.step()
    return pseudo_features
    
\end{lstlisting}

\newpage

\begin{lstlisting}[language=Python, numbers=none]
def pstarc(args, test_loader, netFE, netC):

    netC.train()

    # generate pseudo source features
    pseudo_feats = generate_features(args, netC, num_features=20, num_epochs=50, feature_dim= 256)
    pseudo_scores = nn.Softmax(dim=1)(netC(pseudo_feats))
    pseudo_maxprobs, pseudo_label_bank = torch.max(pseudo_scores, dim=1)
   
    fea_bank = pseudo_feats.cpu()
    fea_bank = torch.nn.functional.normalize(fea_bank)
    score_bank = pseudo_scores
    label_bank = pseudo_label_bank.cpu()

    optimizer = optim.SGD(netC.parameters(), lr =5e-4, momentum = 0.9, weight_decay = 0, nesterov = True)

    iter_test = iter(loader)
    
    for i in range(len(loader)):
        inputs, labels = next(iter_test)
        # Get image and its strong augmentation 
        image, image_s = inputs   

        netFE.train()

        # get image features and its prediction vectors
        features = netFE(image)
        probs_image = nn.Softmax(dim=1)(netC(features))

        # get strong augmented image features and its prediction vectors
        features_s = netFE(image_s)
        probs_image_s = nn.Softmax(dim=1)(features_s)

        # compute sample-wise entropy 
        ent_batch = Entropy(p_image)   
        # get dynamic threshold to select low entropy samples
        ent_thresh = torch.mean(ent_batch)      

        # Computer L_aug
        loss_aug = - torch.sum(probs_image * probs_image_s, dim=1)

        with torch.no_grad():
            f_norm = torch.nn.functional.normalize(features)

            score_near = torch.zeros(image.shape[0], K, class_num).cuda() 
            score_near_cls = torch.zeros(image.shape[0], K, class_num).cuda() 
            for c in range(args.class_num):
                # get pseudo source features and current batch features belonging to class c
                src_cls_feats = torch.nn.functional.normalize(fea_bank[label_bank==c])
                src_cls_scores = score_bank[label_bank==cls_idx]
                curr_cls_feats = output_f_[pseudo_label.cpu()==cls_idx]
                curr_cls_probs = max_prob[pseudo_label.cpu()==cls_idx]

                # Retrieve top K pseudo source features for each test sample
                cos_sim = curr_cls_feats @ src_cls_feats.T                
                cls_dist_near, cls_idx_near = torch.topk(cls_dist, dim=-1, largest=True, k=K + 1)
                cls_dist_near, cls_idx_near = cls_dist_near[:, 1:], cls_idx_near[:, 1:] 
                cls_score_near = src_cls_scores[cls_idx_near]
                score_near_cls[pseudo_label.cpu()==cls_idx] = cls_score_near
                
            # select pseudo source samples as positive for low entropy samples 
            score_near[ent_batch<ent_thresh] = score_near_cls[ent_batch<ent_thresh]
            # self anchor the low entropy samples 
            score_near[ent_batch>ent_thresh] = (probs_image[ent_batch>ent_thresh]).detach().clone().unsqueeze(1).expand(-1, args.K, -1)

        # repeat probs_image(of dimension batch x C) K times
        probs_image_un = probs_image.unsqueeze(1).expand(-1, args.K, -1)  # batch x K x C

        # Computer the attraction loss L_attr 
        loss_postive = -(probs_image_un * score_near).sum(-1).sum(1)

        # Computer the dispersion loss L_attr
        mask = torch.ones((features_w.shape[0], features_w.shape[0]))
        diag_num = torch.diag(mask)
        mask_diag = torch.diag_embed(diag_num)
        mask = mask - mask_diag
        copy = softmax_out.T  
        dot_neg = softmax_out @ copy  # batch x batch
        loss_negative = (dot_neg * mask.cuda()).sum(-1)  # batch

        loss = torch.mean(loss_aug + loss_attr + loss_disp)

        optimizer.zero_grad()
        loss.backward()
        optimizer.step()

        netFE.eval()
        outputs = netC(netFE(image))
        predictions = torch.max(outputs, dim=1)
    
    return 
    
\end{lstlisting}

We will publicly release the code on acceptance. Parts of this code is adapted from AaD.

\section{Augmentation}

We use the same augmentations as used in AdaContrast here. We explicitly report the series of  augmentations done along with their ranges for better reproducibility.

\begin{lstlisting}[language=Python, numbers=none]
from torchvision import transforms

class GaussianBlur(object):
    def __init__(self, sigma=[0.1, 2.0]):
        self.sigma = sigma

    def __call__(self, x):
        sigma = random.uniform(self.sigma[0], self.sigma[1])
        x = x.filter(ImageFilter.GaussianBlur(radius=sigma))
        return x
        
transform_list = [
    transforms.RandomResizedCrop(crop_size, scale=(0.2, 1.0)),
    transforms.RandomApply(
    [transforms.ColorJitter(0.4, 0.4, 0.4, 0.1)],
    p=0.8),
    transforms.RandomGrayscale(p=0.2),
    transforms.RandomApply([GaussianBlur([0.1, 2.0])], p=0.5),
    transforms.RandomHorizontalFlip(),
    transforms.ToTensor()
]
\end{lstlisting}

%% file: spiderweb.tex
\newcommand{\D}{6} % number of dimensions (config option)
\newcommand{\U}{100} % number of scale units (config option)
\newdimen\R % maximal diagram radius (config option)
\R=2.7cm 
\newdimen\L % radius to put dimension labels (config option)
\L=3.1cm
\newcommand{\A}{360/\D} % calculated angle between dimension axes  

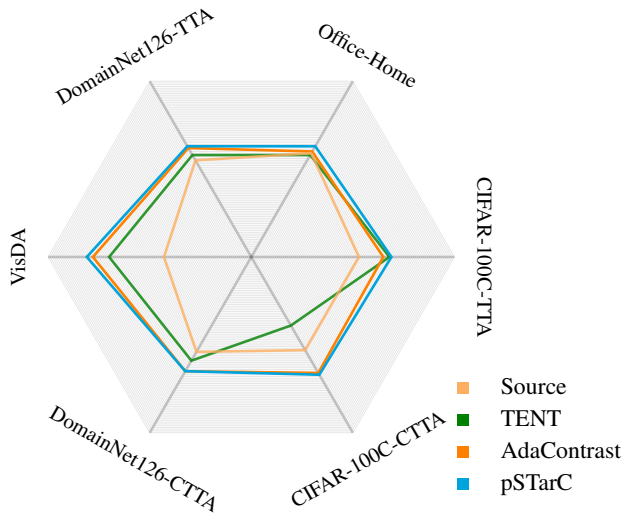
\begin{figure}[t!]
 \centering

\begin{tikzpicture}[scale=1]
  \path (0:0cm) coordinate (O); % define coordinate for origin

  % draw the spiderweb
  \foreach \X in {1,...,\D}{
    \draw [line width=1pt, opacity=0.2] (\X*\A:0) -- (\X*\A:\R);
  }

  \foreach \Y in {0,...,\U}{
    \foreach \X in {1,...,\D}{
      \path (\X*\A:\Y*\R/\U) coordinate (D\X-\Y);
      % \fill (D\X-\Y) circle (1pt);
    }
    \draw [opacity=0.1] (0:\Y*\R/\U) \foreach \X in {1,...,\D}{
        -- (\X*\A:\Y*\R/\U)
    } -- cycle;
  }

  % define labels for each dimension axis (names config option)
  \path (1*\A:\L) node (L1) {\footnotesize \rotatebox{330}{Office-Home}};
  \path (2*\A:\L) node (L2) {\footnotesize \rotatebox{30}{DomainNet126-TTA}};
  \path (3*\A:\L) node (L3) {\footnotesize \rotatebox{90}{VisDA}};
  \path (4*\A:\L) node (L4) {\footnotesize \rotatebox{330}{DomainNet126-CTTA}};
  \path (5*\A:\L) node (L5) {\footnotesize \rotatebox{30}{CIFAR-100C-CTTA}};
  \path (6*\A:\L) node (L6) {\footnotesize \rotatebox{270}{CIFAR-100C-TTA}};

  \draw [color=black!50!green,line width=1pt,opacity=0.8]
    (D1-58.2) --
    (D2-58.9) --
    (D3-70.7) -- 
    (D4-59.7) --
    (D5-39.1) --
    (D6-68.8) -- cycle;

  \draw [color=orange,line width=1pt,opacity=1.0]
    (D1-60.2) --
    (D2-62.6) --
    (D3-78.7) -- 
    (D4-65.1) --
    (D5-66.6) --
    (D6-65.9) -- cycle;

  \draw [color=cyan!90!black,line width=1pt,opacity=1.0]
    (D1-63.5) --
    (D2-63.7) --
    (D3-81.9) -- 
    (D4-65.5) --
    (D5-67.7) --
    (D6-69.5) --cycle;

  \draw [color=yellow1,line width=1pt,opacity=0.8]
    (D1-59.4) --
    (D2-55.2) --
    (D3-43.8) --
    (D4-54.7) --
    (D5-53.6) --
    (D6-53.6) -- cycle;

\newcommand\ColorBox[1]{\textcolor{#1}{\rule{1ex}{1ex}}}
\node[anchor=south west,xshift=-20pt,yshift=5pt] at (current bounding box.south east) 
{
\begin{tabular}{@{}lp{3cm}@{}}
\ColorBox{yellow1!}& {\small Source          }\\
\ColorBox{black!50!green!}   & {\small TENT          }\\
\ColorBox{orange!} & {\small AdaContrast   }\\
\ColorBox{cyan!90!black!}  & {\small pSTarC        }\\
\end{tabular}
};

\end{tikzpicture}
\caption{Overall comparison of pSTarC with TTA methods.}
\label{fig:spiderweb}
\end{figure}